\documentclass[11pt,a4paper]{article}
\usepackage[hyperref]{emnlp-ijcnlp-2019}
\usepackage{times}
\usepackage{latexsym}

\usepackage{url}

\usepackage{microtype}
\usepackage{graphicx}
\usepackage{subfigure}
\usepackage{amsmath}
\usepackage{amsfonts}
\usepackage{booktabs} 

\aclfinalcopy % Uncomment this line for the final submission

\title{Decomposing Textual Information For Style Transfer}

\author{Ivan P. Yamshchikov\footnotemark[1]  \\
  Max Planck Institute \\
  for Mathematics in the Sciences \\
  Leipzig, Germany \\
  \texttt{ivan@yamshchikov.info} \\\And
  Viacheslav Shibaev  \\
  Ural Federal University \\
  Ekaterinburg, Russia \\\And
  Aleksander Nagaev \\
  Ural Federal University \\
  Ekaterinburg, Russia \\\AND
  J\"urgen Jost  \\
  Max Planck Institute \\
  for Mathematics in the Sciences \\
  Leipzig, Germany \\\And
  Alexey Tikhonov\thanks{Equal contribution} \\
  Yandex \\
  Berlin, Germany \\
  \texttt{altsoph@gmail.com} \\}

\date{}

\begin{document}
\maketitle
\begin{abstract}
This paper focuses on latent representations that could effectively decompose different aspects of textual information. Using a framework of style transfer for texts, we propose several empirical methods to assess information decomposition quality. We validate these methods with several state-of-the-art textual style transfer methods. Higher quality of information decomposition corresponds to higher performance in terms of bilingual evaluation understudy (BLEU) between output and human-written reformulations.
\end{abstract}

\section{Introduction}
\label{intro}

% Significant progress in many classical machine vision problems is achieved with autoencoders \cite{kingma13} and generative adversarial networks \cite{goodfellow}. One can name several examples, such as style transfer for images \cite{Gatys}, image generation \cite{Radford} and learning interpretable image representations \cite{Chen}. Indeed, 

The arrival of deep learning seems transformative for many areas of information processing and is especially interesting for generative models \cite{hu17}. However, natural language generation is still a challenging task due to a number of factors that include the absence of local information continuity and non-smooth disentangled representations \cite{bowman}, and discrete nature of textual information \cite{hylsx}.  If information needed for different natural language processing (NLP) tasks could be encapsulated in independent components of the obtained latent representations, one could have worked with different aspects of text independently. This could also naturally simplify learning transfer for NLP models and potentially make them more interpretable. 

Despite the fact that content and style are deeply fused in natural language, style transfer for texts is often addressed in the context of disentangled latent representations \cite{hylsx, Shen, fu2, john18, romanov18, tian18}. A majority of these works use an encoder-decoder architecture with one or multiple style discriminators to improve latent representations. An encoder takes a given sentence as an input and generates a style-independent content representation. The decoder then uses this content representation and a target style representation to generate a new sentence in the needed style. This approach seems intuitive and appealing but has certain difficulties. For example, \citet{subramanian18} question the quality and usability of the disentangled representations for texts with an elegant experiment. The authors train a state of the art architecture that relies on disentangled representations and show that an external artificial neural network can predict the style of the input using a semantic component of an obtained latent representation (that supposedly did not incorporate stylistic information). 

In this work, we demonstrate that the decomposition of latent representations is, indeed, attainable with encoder-decoder based methods but depends on the used architecture. Moreover, architectures with higher quality of information decomposition perform better in terms of the style transfer task.

The contribution of this paper is threefold: (1) we propose several ways to quantify the quality of the obtained latent semantic representations; (2) we show that the quality of such representation can significantly differ depending on the used architecture; (3) finally we demonstrate that architectures with higher quality of information decomposition perform better in terms of BLEU \cite{Papineni} between output of a model and a human written reformulations. 

\section{Related Work}
\label{sec:rw}
 
It is hard to define style transfer rigorously \cite{Xu3}. Therefore recent contributions in the field are mostly motivated by several empirical results and rather address specific narrow aspects of style that could be empirically measured. Stylistic attributes of text include author-specific attributes (see \cite{xu} or \cite{Jhamtani} on 'shakespearization'), politeness \cite{Sennrich}, the \textit{'style of the time'} \cite{Hughes}, gender or political slant \cite{Prabhumoye}, and formality of speech \cite{Rao}. All these attributes are defined with varying degrees of rigor. Meanwhile, the general notion of literally style is only addressed in a very broad context. For example, \citet{Hughes} shows that the style of a text can be characterized quantitively and not only with an expert opinion; \citet{Potash} demonstrate that stylized texts could be generated if a system is trained on a dataset of stylistically similar texts; and literary styles of the authors could be learned end-to-end \cite{TY, TYwilde, Vechtomova}. 

In this particular submission we focus on a very narrow framework of sentiment transfer. There is certain controversy whether sentiment of a text could be regarded as its stylistic attribute, see \cite{TYwrong}. However, there seems to be certain agreement in the field that sentiment could be regarded as a viable attribute to be changed by the style transfer system. Addressing the problem of sentiment transfer \citet{Kabbara, li, Xu2} estimate the quality of the style transfer with a pre-trained binary sentiment classifier. \citet{fu2} and \citet{Ficler} generalize this ad-hoc approach and in principle enable the information decomposition approach. They define a style as a set of arbitrary quantitatively measurable categorical or continuous parameters that could be automatically estimated with an external classifier. In this submission we stay within this empirical paradigm of literary style.

Generally speaking, a solution that works for one aspect of a style could not be applied for a different aspect of it. For example, a retrieve-edit approach by \cite{guu} works for sentiment transfer.  A  delete-retrieve model shows good results for sentiment transfer in \cite{li}. However, these retrieval approaches could hardly be used for the style of the time or formality or any other case when the system is expected to paraphrase a given sentence to achieve the target style. To address this challenge \citet{hylsx} propose a more general approach to the controlled text generation combining variational autoencoder (VAE) with an extended wake-sleep mechanism in which the sleep procedure updates both the generator and external discriminator that assesses generated samples and feedbacks learning signals to the generator.  Labels for style were concatenated with the text representation of the encoder and used with "hard-coded" information about the sentiment of the output as the input of the decoder. This approach is promising and is used in many recent contributions. \citet{Shen} use an adversarial loss to decompose information about the form of a sentence and apply a GAN to align hidden representations of sentences from two corpora. \citet{fu2} use an adversarial network to make sure that the output of the encoder does not include stylistic information. \citet{hylsx} also use an adversarial component to ensure there is no stylistic information within the representation. A dedicated component that controls semantic component of the latent representation is proposed by \citet{john18} who demonstrate that decomposition of style and content could be improved with an auxiliary multi-task for label prediction and adversarial objective for a bag-of-words prediction. \citet{romanov18} also introduce a dedicated component to control semantic aspects of latent representations and an adversarial-motivational training that includes a special motivational loss to encourage a better decomposition. 

The framework of information decomposition within latent representations is challenged by an alternative family of neural machine translation approaches. These are works on style transfer with \cite{Carlson} and without parallel corpora \cite{zhang18} in line with \cite{Lample} and \cite{Artetxe}. In particular, \citet{subramanian18} state that learning a latent representation, which is independent of the attributes specifying its style is rarely attainable. They experiment with the model developed in \cite{fu2} where by design the discriminator, which was trained adversarially and jointly with the model, gets worse at predicting the sentiment of the input when the coefficient of the adversarial loss increases. Authors show that a classifier that is separately trained on the resulting encoder representations easily recovers the sentiment of a latent representation produced by the encoder.

In this paper, we show that contrary to \cite{subramanian18} decomposition of the stylistic and semantic information is attainable with autoencoder-type models and could be quantified. However, the quality of such decomposition severely depends on the particular architecture. We propose three different measures for information decomposition quality and using four different architectures show that models with better information decomposition outperform the state-of-the-art models in terms of BLEU between output and human-written reformulations.

%%%%%%%%%%%%%%%%%%%%%%%%%%%%%%%%%%%%%%%%%%%%%%%%%%%%%%%%%%%%%%%%%%%%%%%%%%%%%%%%%%%%%%%%%%%%%%%%%%%%%%%
\section{Style transfer}
\label{sec:st}
\begin{figure}[ht]
\begin{center}
\centerline{\includegraphics[width=\columnwidth]{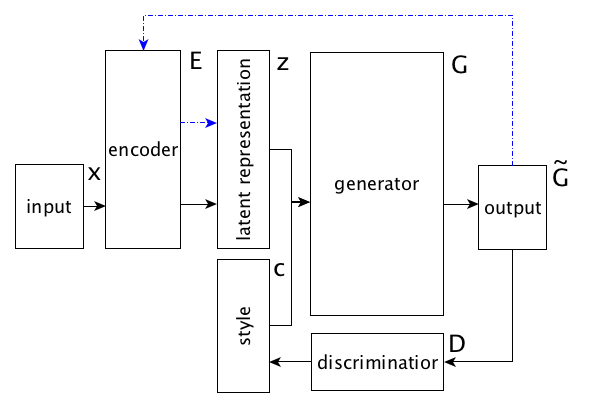}}
\caption{The generative model, where style is a structured code targeting sentence attributes to control.
Blue dashed arrows denote the proposed independence constraint of latent representation and controlled attribute, see \cite{hylsx} for the details.}
\label{pic:hu}
\end{center}
\end{figure}

In this work we experiment with extensions of a model, described in \cite{hylsx}, using Texar \cite{hutexar} framework. To generate plausible sentences with specific semantic and stylistic features every sentence is conditioned on a representation vector $z$ which is concatenated with a particular code $c$ that specifies desired attribute, see Figure \ref{pic:hu}. Under notation introduced in \cite{hylsx} the base autoencoder (AE) includes a conditional probabilistic encoder $E$ defined with parameters $\theta_E$ to infer the latent representation $z$ given input $x$
$$z \sim E(x) = q_{E}(z,c|x).$$
Generator $G$ defined with parameters  $\theta_G$ is a GRU-RNN for generating and output $\hat{x}$ defined as a sequence of tokens $\hat{x} = {\hat{x}_1, ..., \hat{x}_T}$ conditioned on the latent representation $z$ and a stylistic component $c$ that are concatenated and give rise to a generative distribution
$$\hat{x} \sim G(z,c) = p_G(\hat{x}|z, c).$$
These encoder and generator form an AE with the following loss
\begin{equation}
\label{eq:lossae}
\mathcal{L}_{ae} (\theta_G, \theta_E; x,c) = - \mathbb{E}_{q_{E} (z,c|x)} \left[ \log q_G (x|z, c) \right].
\end{equation}

This standard reconstruction loss that drives the generator to produce realistic sentences is combined with two additional losses. The first discriminator provides extra learning signals which enforce the generator to produce coherent attributes that match the structured code in $c$. Since it is impossible to propagate gradients from the discriminator through the discrete sample $\hat{x}$, we use a deterministic continuous approximation a "soft" generated sentence, denoted as $\tilde{G} = \tilde{G}_\tau (z, c)$ with "temperature" $\tau$ set to $\tau \rightarrow 0$ as training proceeds. The resulting “soft” generated sentence is fed into the discriminator to measure the fitness to the target attribute, leading to the following loss
\begin{equation}
\mathcal{L}_{c} (\theta_G, \theta_E; x) = -\mathbb{E}_{q_{E} (z,c|x)}  \left[ \log q_D (c | \tilde{G}) \right]. \label{eq:lossc}
\end{equation}

Finally, under the assumption that each structured attribute of generated sentences is controlled through the corresponding code in $c$ and is independent from $z$ one would like to control that other not explicitly modelled attributes do not entangle with  $c$. This is addressed by the  dedicated loss
\begin{equation}
\label{eq:lossz}
\mathcal{L}_{z} (\theta_G; x) = - \mathbb{E}_{q_{E} (z,c|x) q_{D} (c|x)} \left[ \log q_E (z | \tilde{G}) \right].
\end{equation}
The training objective for the baseline, shown in Figure \ref{pic:hu}, is therefore a sum of the losses from Equations (\ref{eq:lossae}) -- (\ref{eq:lossz}) defined as
\begin{equation}
\label{eq:genhu}
min_{\theta_G} \mathcal{L}_{baseline} = \mathcal{L}_{ae} + \lambda_c \mathcal{L}_{c} + \lambda_z \mathcal{L}_{z},
\end{equation}
where $\lambda_c$ and $\lambda_z$ are balancing parameters.

\begin{figure}[ht]
\begin{center}
\centerline{\includegraphics[width=\columnwidth]{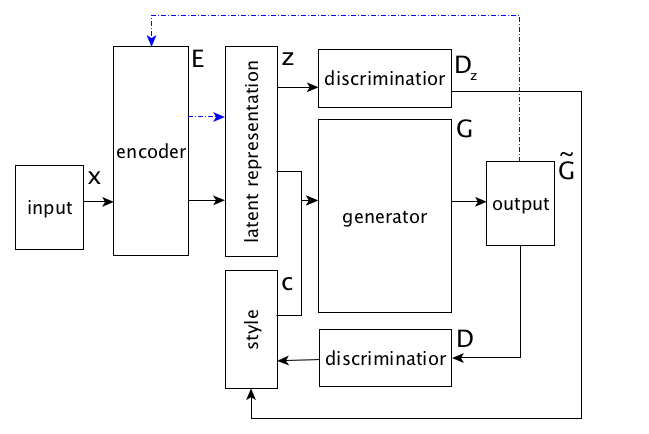}}
\caption{The generative model with dedicated discriminator introduced to ensure that semantic part of the latent representation does not have information on the style of the text.}
\label{pic:d}
\end{center}
\end{figure}

Let us propose two further extensions of this baseline architecture. To improve reproducibility of the research the code of the studied models is open\footnote{https://github.com/VAShibaev/textstyletransfer}. Both extensions aim to improve the quality of information decomposition within the latent representation. In the first one, shown in Figure \ref{pic:d}, a special dedicated discriminator is added to the model to control that the latent representation does not contain stylistic information. The loss of this discriminator is defined as
\begin{equation}
\label{eq:lossdisc}
\mathcal{L}_{D_z} (\theta_G; x,c) = - \mathbb{E}_{q_{E} (z|x)} \left[ \log q_{D_z} (c | z) \right].
\end{equation}

Here a discriminator denoted as $D_z$ is trying to predict code $c$ using representation $z$. Combining the loss defined by Equation (\ref{eq:genhu}) with the adversarial component defined in Equation (\ref{eq:lossdisc}) the following learning objective is formed
\begin{equation}
\label{eq:gendz}
min_{\theta_G} \mathcal{L} = \mathcal{L}_{baseline} - \lambda_{D_z} \mathcal{L}_{Dz},
\end{equation}
where $\mathcal{L}_{baseline}$ is a sum defined in Equation (\ref{eq:genhu}), $\lambda_{D_z}$ is a balancing parameter.

\begin{figure}[ht]
\begin{center}
\centerline{\includegraphics[width=\columnwidth]{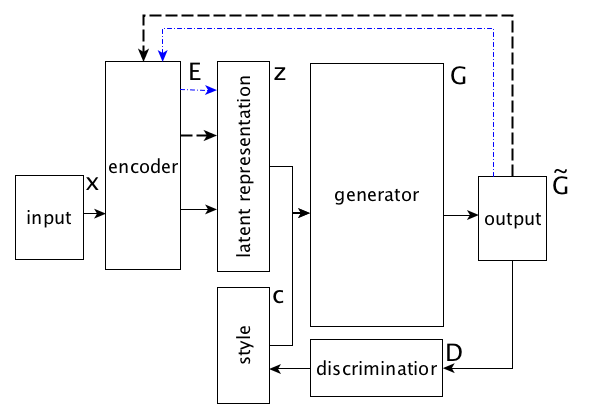}}
\caption{The generative model with a dedicated loss added to control that semantic representation of the output, when processed by the encoder, is close to the semantic representation of the input.}
\label{pic:sae}
\end{center}
\end{figure}

The second extension of the baseline architecture does not use an adversarial component $D_z$ that is trying to eradicate information on $c$ from component $z$. Instead, the system, shown in Figure \ref{pic:sae} feeds the "soft" generated sentence $\tilde{G}$ into encoder $E$ and checks how close is the representation $E(\tilde{G} )$ to the original representation $z = E(x)$ in terms of the cosine distance. We further refer to it as {\em shifted autoencoder} or SAE. Ideally, both $E(\tilde{G} (E(x), c))$ and $E(\tilde{G} (E(x), \bar{c}))$, where $\bar{c}$ denotes an inverse style code, should be both equal to $E(x)$\footnote{This notation is valid under the assumption that every stylistic attribute is a binary feature}. The loss of the shifted autoencoder is 
\begin{equation}
\label{eq:gesae}
min_{\theta_G} \mathcal{L} = \mathcal{L}_{baseline} +  \lambda_{cos} \mathcal{L}_{cos} +  \lambda_{cos^{-}} \mathcal{L}_{cos^{-}},
\end{equation}
where $\lambda_{cos}$ and $\lambda_{cos^{-}}$ are two balancing parameters, with two additional terms in the loss, namely, cosine distances between the softened output processed by the encoder and the encoded original input, defined as 
\begin{eqnarray}
\label{eq:cosloss}
\mathcal{L}_{cos} (x,c) = \cos \left( E(\tilde{G}(E(x), c)), E(x) \right),  \nonumber \\
\mathcal{L}_{cos^{-}} (x,c) = \cos \left( E(\tilde{G}(E(x), \bar{c})), E(x) \right).
\end{eqnarray}

\begin{figure}[ht]
\begin{center}
\centerline{\includegraphics[width=\columnwidth]{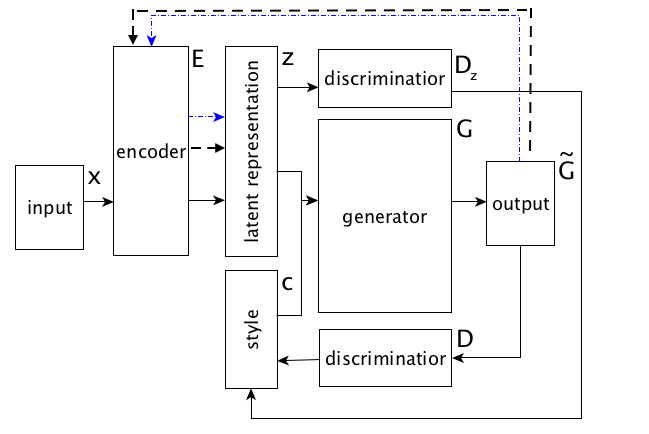}}
\caption{A combination of an additional discriminator used in Figure \ref{pic:d} with a shifted autoencoder shown in Figure \ref{pic:sae}}
\label{pic:combo}
\end{center}
\end{figure}

We also study a combination of both approaches described above, shown on Figure \ref{pic:combo}.

\citet{TSSNY} carry out a series of experiments for these architectures. In this contribution, we work with the same data set of human-labeled positive and negative reviews but focus solely on the quality of information decomposition. 

\section{Information decomposition for texts}
\label{sec:de}

As we have mentioned earlier, several recent contributions rely on the idea that decomposing different aspects of textual information into various components of a latent representation might be helpful for a task of style transfer. To our knowledge, this is a supposition that is rarely addressed rigorously. The majority of the arguments in favor of information decomposition based architectures is of an intuitive and qualitative rather than quantitative nature. Moreover, there are specific arguments against this idea. 

In particular, \citet{subramanian18} show that information decomposition does not necessarily occur in autoencoder-based systems using a method developed in \cite{fu2}. \citet{subramanian18} demonstrate that as training proceeds, the internal discriminator, which was trained adversarially and jointly with the model, gets worse at predicting the sentiment of the input. However, an external classifier that is separately trained on the resulting latent representations easily recovers the sentiment. This is a strong argument in favor of the idea that actual disentanglement does not happen. Instead of decomposing the semantic and stylistic aspects of information, the encoder merely 'tricks' internal classifier and 'hides' stylistic information in the semantic component ending up in some local optimum.  

\subsection{Empirical measure of information decomposition quality}
\label{sec:em}

Yelp!\footnote{https://www.yelp.com/dataset} reviews dataset that was lately enhanced with human written reformulations by \cite{tian18} is one of the most frequently used baselines for textual style transfer at the moment. It consists of restaurant reviews split into two categories, namely, positive and negative. There is a human written reformulation of every review in which the sentiment is changed that is commonly used as a ground truth for the task performance estimation. 

\begin{figure}[ht]
\begin{center}
\centerline{\includegraphics[width=\columnwidth]{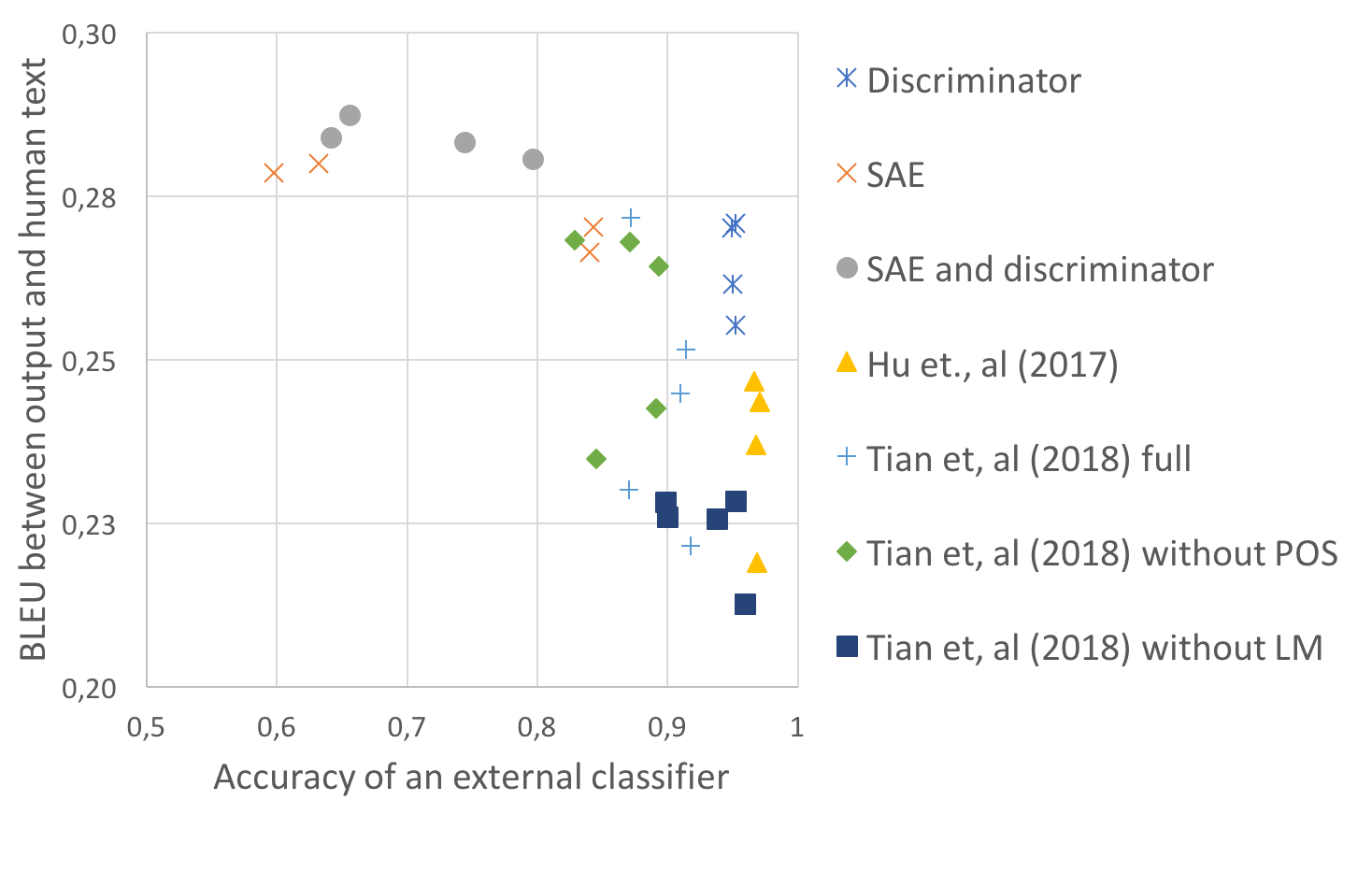}}
\caption{BLEU between system's output and human-written reformulations seems to be higher if accuracy of an external classifier is closer to one half. Systems that decompose information better tend to show higher BLEU.}
\label{pic:bleudecomp}
\end{center}
\end{figure}

We applied an empirical method to estimate the quality of information decomposition to the architectures described in Section \ref{sec:st} as well as architectures developed by \cite{tian18}. An external classifier was trained from scratch to predict a style of a message using component $z$ of a latent representation produced by an encoder. If information decomposition does not happen, one would expect that accuracy of an external classifier would be close to $1$. This would mean that despite intuitive expectations, information about the style of a message is present in $z$. If decomposition were effective, the accuracy of an external classifier would be close to $0.5$; in \cite{TSSNY} it is shown that style transfer methods show varying results in terms of accuracy and BLEU for different retrains, so in this paper the accuracy of an external classifier and BLEU between the system's output and human-written reformulations was measured after four independent retrains. On Figure \ref{pic:bleudecomp}, one can see the results of these experiments.

The fact that the external classifier always predicts style with the probability that is above one half could be partially attributed to the fact that full information decomposition of sentiment and semantics is hardly attainable.  For example, such adjectives as "delicious" or "yummy" incorporate positive sentiment with the semantics of taste, whereas "polite" or "friendly" in Yelp! reviews are combining positive sentiment with the semantics of service. This internal entanglement of sentiment and semantics is discussed in detail in \cite{TYwrong}. It is essential to mention that the very fact that semantics and stylistics are entangled on the level of words does not deny a theoretical possibility to build a latent representation where they are fully disentangled. Anyway, Figure \ref{pic:bleudecomp} demonstrates that the quality of the disentanglement is much better for SAE-type architectures. Since the shifted autoencoder controls the cosine distance between soft output and input, the encoder has to disentangle the semantic component, rather than "hide" the sentiment information from the discriminator.

\begin{figure}[ht]
\begin{center}
\centerline{\includegraphics[width=\columnwidth]{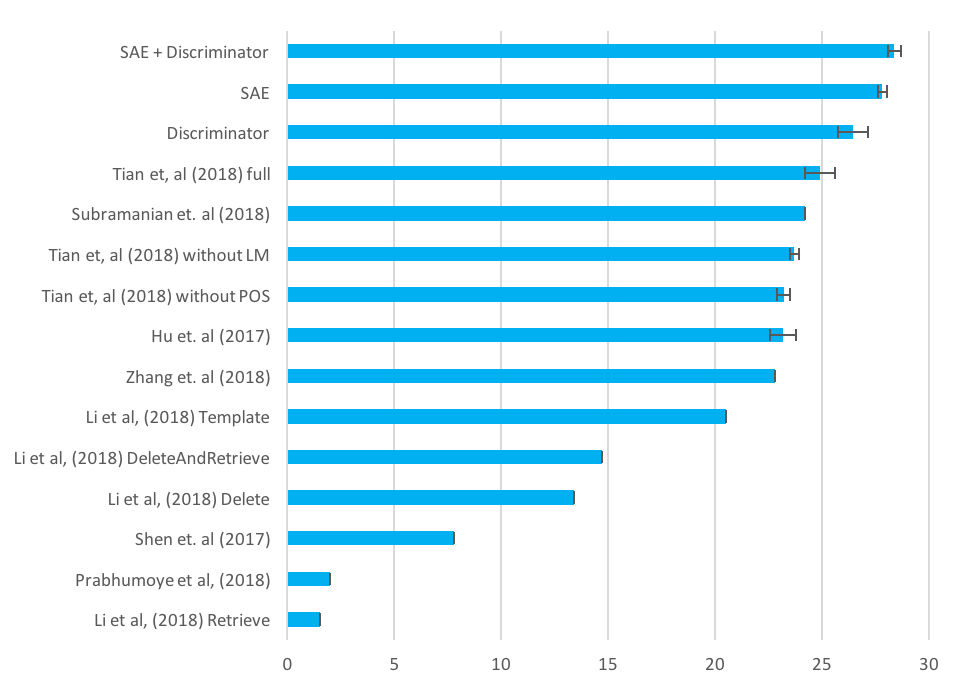}}
\caption{Overview of the BLEU between output and human-written reformulations of Yelp! reviews. Architecture with additional discriminator, shifted autoencoder (SAE) with additional cosine losses, and a combination of these two architectures measured after five re-runs outperform the baseline by \cite{hylsx} as well as other state of the art models. Results of \cite{romanov18} are not displayed due to the absence of self-reported BLEU scores}
\label{pic:ovreal}
\end{center}
\end{figure}

On Figure \ref{pic:ovreal} one can see how state of the art approaches compare to each other in terms of BLEU between output and human-written reformulations. All systems were retrained five times from scratch to report error margins of the methods since the results are noisy. BLEU between output and human-written reformulations is higher for lower values of external classifier accuracy. Systems that perform better in terms of information decomposition outperform system with lower quality of information decomposition. Moreover, the system that does not rely on an idea of disentangled latent representations at all shows weaker results than systems with high information disentanglement. It is important to note that there is a variety of methods to assess the quality of style transfer such as PINC (Paraphrase In N-gram Changes) score \cite{Carlson}, POS distance \cite{tian18}, language fluency \cite{john18}, etc. The methodology of style transfer quality assessment is addressed in detail in \cite{TSSNY}, but BLEU between output and input is a very natural all-purpose metric for the task of such type that is common in the style transfer literature.  

Tables \ref{tab:ex} - \ref{tab:ex2} allow to compare random examples for different architectures. Generally, baseline and discriminator perform poorly once the syntax of a review is irregular or if there are some omissions in the text. SAE-based architectures tend to preserve the semantic component better. They also add sentimentally charged words at random not as often as the baseline and the discriminator-based architecture.

\begin{table*}[t!]
\centering
\small{\begin{tabular}{lll}
  input	& Human	& baseline \\
  \hline
  the carne asada burrito is awesome! &	the carne asada burrito is awful! &	the worst asada burrito is gross!\\
      \hline
    the rooms are not that nice and &	the rooms were spacious and	 & the rooms are excellent that nice and \\
        the food is not that good either. & food was very well cooked	 & the food is not that good either.\\
        \hline
    it was so delicious; &	everything tasted bad,  &	it was so rude; \\
     i've never had anything like it! & nothing i liked & i've never had anything like it! \\
     \hline
so, that was my one and only &	i will be ordering the  &	so, that was my one and best \\
time ordering the benedict there. & benedict again very good meal! & time ordering the perfect there. \\
\hline
you'll see why once you get there. &	you'll see why i don't like it &	you'll see why once you get there. \\
& once you get there. & \\
\hline
i wanted to like this place but &	a place like this &	i helped to like this place \\
 it just became a big disappointment. & is a great value & it just became a big hidden. \\
 \hline
and i had my sugar bowl favorite, &	my sugar bowl favorite &	and i had my sugar bowl worst, \\
the top hat sundae! & was not in stock. & the lackluster hat gross! \\
\hline
um... we just told him that & um... we just told him, &	amazing... we just told him that  \\
 we didn't want to finance.	& sure we'd go ahead and finance! & we did definitely want to open. \\
\hline
definitely a place to keep in mind. &	not a place i would recommend &	disappointing a place to keep in mind. \\
\hline
firstly, their fees are generally &	the fees are comparable &	best, their fees are generally  \\
higher than other places. & to other places. & higher than other places. \\
\hline
love the afternoon - &	hate the aternoon & absurd the inappropriate -  \\
tea at the phoenician. & tea at the phoenician & tea at the insult. \\
\end{tabular}}
\caption{Several random input lines alongside with human written reformulation and the reformulation generated by the baseline.}
  \label{tab:ex}
\end{table*}

\begin{table*}[t!]
\centering
\small{\begin{tabular}{lll}
   Discriminator & SAE & SAE + Discriminator \\
  \hline
  the carne asada burrito is absurd! & the carne asada burrito is worst! &	the carne asada burrito is sub-par! \\
  \hline
the rooms are delicious that nice &	the rooms are definitely that nice and &	the rooms are consistantly that nice and\\
and the food is delicious that good either. & the food is definitely that good either. & the food is consistantly that good either. \\
\hline
it was so not; i've never &	it was so disgusting; i've never &	it was so angry ; i've never \\
had anything like it! & had anything like it! & had anything like it! \\
\hline
so, that was my one and fam &	so, that was my one and kids  &	so, that was my one and always \\
time solid the benedict perfectly. & time ordering the benedict there. &	 time ordering the benedict there. \\
\hline
you'll trash why once you get there. &	you'll avoid why once you get there. &	you'll see why once you get there. \\
\hline
i wanted to like this place but  &	i wanted to like this place but &	i wanted to like this place but \\
it just delightful a big genius. &	 it just became a big midwest. & it just mildly a big stocked. \\
\hline
and i had my sugar bowl favorite, &	and i had my sugar bowl broken, &	and i had my sugar bowl misleading, \\
 the absurd hurts ache! &	 the garage hat holes! & the top quesadilla sundae! \\
\hline
expertly... we just delightful him &	um... we just loved him that &	um... we just entertained him that  \\
that we did magical want adds marvelous. &	 we did definatly want to finance. &	we did perfected want to incredible. \\
\hline
ridiculous a place to keep in mind. &	would a place to keep in mind. &	wont a place to keep in mind. \\
\hline
firstly, their project are  &	firstly, their draw are  &	sheila, their round are  \\
generally higher than other places. & generally higher than other places. &	generally higher than other places. \\
\hline
horrific the trap - tea at the gut. &	dumb the afternoon - tea at the rabbit. &	wtf the afternoon - tea at the slim.
\end{tabular}}
\caption{Reformulations generated by the baseline with additional discriminator, shifted autoencoder and shifted autoencoder with additional discriminator corresponding to the inputs in Table \ref{tab:ex}.
}
  \label{tab:ex2}
\end{table*}

\subsection{Preservation of semantic component}
\label{sec:psc}

\begin{figure}[ht]
\begin{center}
\centerline{\includegraphics[width=\columnwidth]{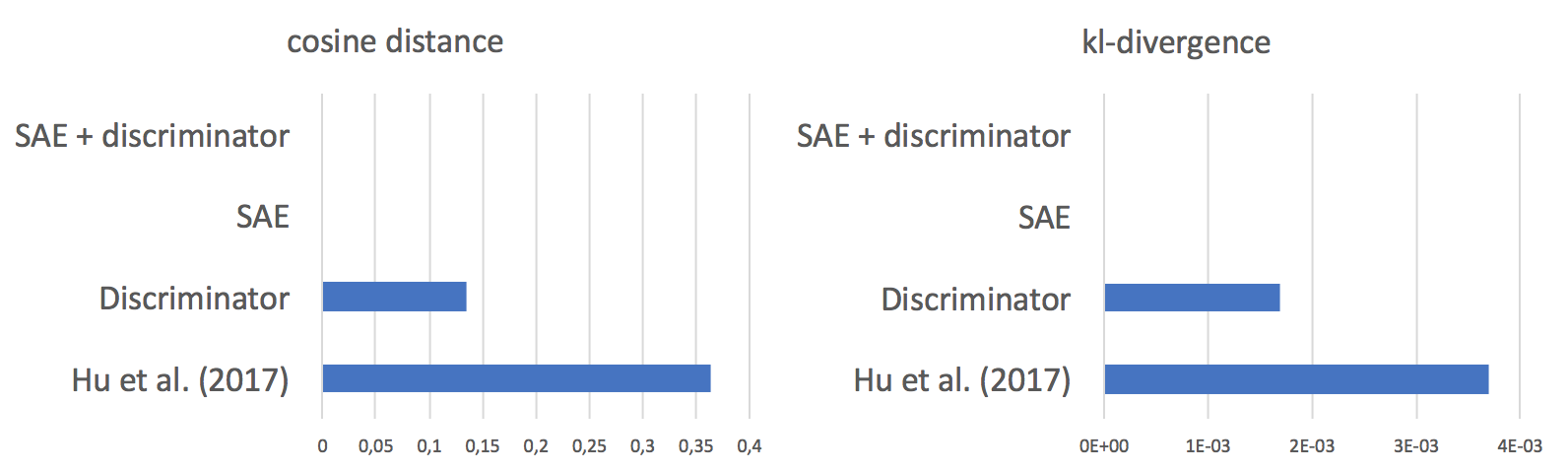}}
\caption{Comparison of cosine distances and KL-divergences between semantic components of latent representation for inputs and outputs. After 12 epochs of training SAE makes semantic component $z$ for every output equal to the semantic component for a corresponding input. Discriminator corresponds to lower values of KL-divergence and cosine distance then baseline \cite{hylsx}}
\label{pic:coskl}
\end{center}
\end{figure}

Another way to quantify the quality of latent representations is to calculate cosine distance and KL-divergence between semantic components of latent representations for the inputs and corresponding outputs. If we believe that the latent representation captures the semantics of the input that should be preserved in the output, the ideal behavior of the system is to produce equal latent representation for both the input and the output phrase. Indeed, on Figure \ref{pic:coskl} one can see that SAE manages to learn a space of latent representations in which semantic components of inputs and outputs are always equal to each other. Architecture with additional stylistic discriminator shows lower cosine distances and lower KL-divergences then the baseline yet. This results are in line with the measurements discussed above in Section \ref{sec:em}.

\begin{figure}[ht]
\begin{center}
\centerline{\includegraphics[width=\columnwidth]{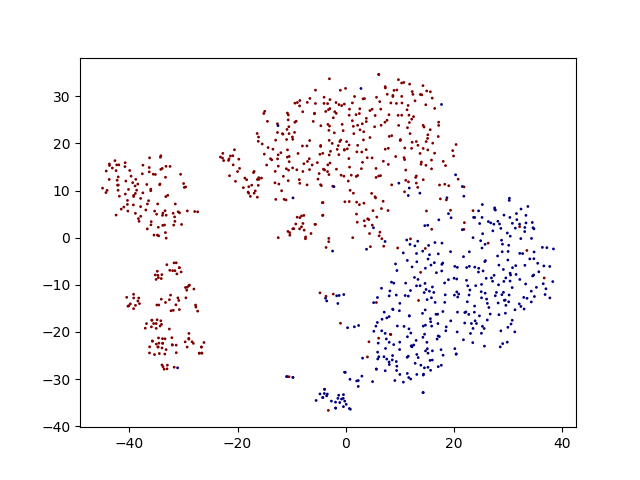}}
\caption{t-SNE visualisation of the obtained latent representations for the baseline architecture proposed in \cite{hylsx}. Red dots represent positive reviews. Blue dots represent negative reviews. One can clearly see that stylistic information can be recovered from the representation.}
\label{pic:baseviz}
\end{center}
\end{figure}

To get an intuition on how the resulting latent space differs for different architectures, one can look at the t-SNE visualizations \cite{tsne} for the resulting latent representations of the data that different systems produce. In Figure \ref{pic:baseviz}, one can see that the baseline latent representations easily allow recovering the sentiment.

\begin{figure}[!ht]
\begin{center}
\centerline{\includegraphics[width=\columnwidth]{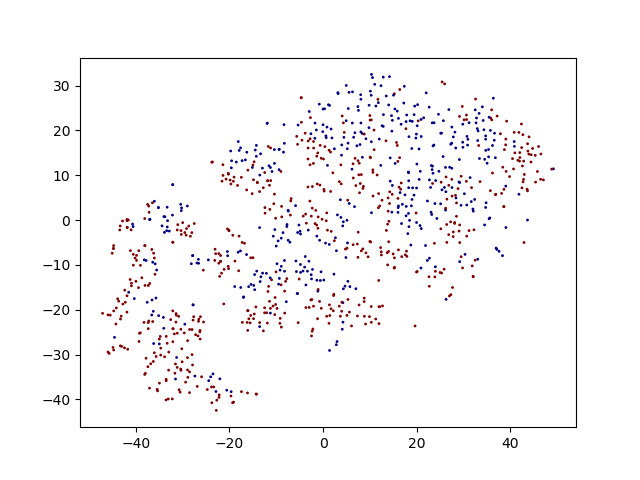}}
\caption{t-SNE visualisation of the obtained latent representations for the architecture with an additional discriminator. Red dots represent positive reviews. Blue dots represent negative reviews. One can see that it is harder to recover stylistic information from the representation.}
\label{pic:discviz}
\end{center}
\end{figure}

In contrast with the baseline, the architecture with additional discriminator obtains better disentanglement. Figure \ref{pic:discviz} shows that in this case one has a harder time recovering the sentiment of the sentence based on its latent representation. 

\begin{figure}[!ht]
\begin{center}
\centerline{\includegraphics[width=\columnwidth]{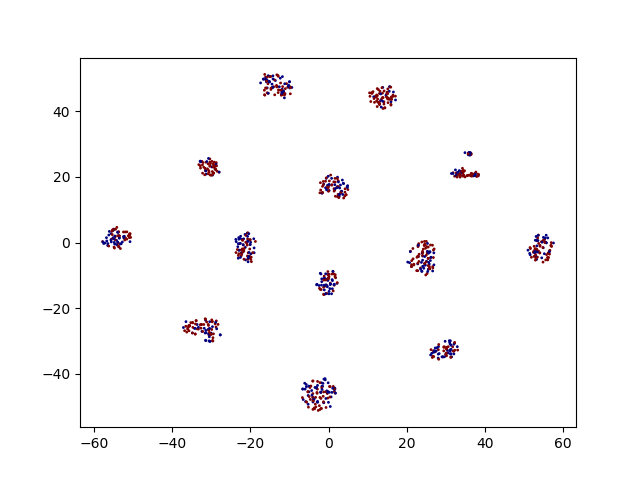}}
\caption{t-SNE visualisation of the obtained latent representations for the shifted autoencoder. Red dots represent positive reviews. Blue dots represent negative reviews. One can see that it is harder to recover stylistic information from the representation and the structure of the differs significantly from the latent representation space obtained by the baseline.}
\label{pic:saevis}
\end{center}
\end{figure}

SAE does not only show a higher level of disentanglement but also produces equal semantic components for the input and the corresponding output. Judging by Figure \ref{pic:saevis} this makes SAE representations denser in certain areas of the semantic space and sparser in the others. 

Aligning results shown on Figures \ref{pic:bleudecomp} - \ref{pic:saevis} one can clearly see several crucial things: (1) architectures based on the idea of disentangled latent representations show varying performance in terms of BLEU between output and human written reformulations; (2) architectures with higher quality of information decomposition in terms of correlation or KL-divergence between representations for input and output, show higher performance; (3) architectures that produce equal semantic components for a given input and corresponding output show the highest performance; (4) these results are aligned with empirical estimation of decomposition quality with external classifiers; it shows that architectures that are more successfully disentangling semantics of the input from its stylistics tend to perform better.

\section{Conclusion}

This paper addresses the questions of information decomposition for the task of textual style transfer. We propose three new architectures that use latent representations to decompose stylistic and semantics information of input. Two different methods to assess the quality of such decomposition are proposed. It is shown that architectures that produce an equal semantic component of latent representations for input and corresponding output outperform state of the art architectures in terms of BLEU between output and human written reformulations. An empirical method to assess the quality of information decomposition is proposed. There is a correspondence between higher BLEU between output and human written reformulations and better quality of information decomposition.

\bibliography{acl2019}
\bibliographystyle{acl_natbib}

\end{document}